\documentclass[final]{cvpr}

\usepackage{times}
\usepackage{epsfig}
\usepackage{graphicx}
\usepackage{amsmath}
\usepackage{amssymb}

\usepackage{array}
\usepackage{color}
\usepackage{booktabs}
\usepackage{multirow}
\usepackage{siunitx}
\usepackage{fixmath}
\usepackage{etoolbox}
\usepackage{algpseudocode}
\usepackage[numbers,sort]{natbib}
\robustify\bfseries

\usepackage[pagebackref=true,breaklinks=true,colorlinks]{hyperref}

\newcommand{\PAR}[1]{\smallskip \noindent \textbf{#1}}
\newcommand{\PARbegin}[1]{\noindent \textbf{#1}}

\renewcommand{\figref}[1]{Figure~\ref{#1}}
\renewcommand{\tabref}[1]{Table~\ref{#1}}
\renewcommand{\eqref}[1]{Eq.~(\ref{#1})}

\renewcommand{\vec}[1]{\mathbold{#1}}
\newcommand{\mat}[1]{\mathbold{#1}}

\begin{document}

\title{Layout-Guided Novel View Synthesis from a Single Indoor Panorama}

\author{
Jiale Xu\textsuperscript{\rm 1} \quad
Jia Zheng\textsuperscript{\rm 2} \quad
Yanyu Xu\textsuperscript{\rm 3} \quad
Rui Tang\textsuperscript{\rm 2} \quad
Shenghua Gao\textsuperscript{\rm 1,4}\thanks{Corresponding author.} \\
\textsuperscript{\rm 1}ShanghaiTech University \quad
\textsuperscript{\rm 2}KooLab, Manycore \\
\textsuperscript{\rm 3}Institute of High Performance Computing, A*STAR \\
\textsuperscript{\rm 4}Shanghai Engineering Research Center of Intelligent Vision and Imaging \\
{\tt\small \{xujl1, gaoshh\}@shanghaitech.edu.cn} \quad
{\tt\small \{jiajia, ati\}@qunhemail.com} \\
{\tt\small xu\_yanyu@ihpc.a-star.edu.sg}
}

\maketitle

\begin{abstract}
Existing view synthesis methods mainly focus on the perspective images and have shown promising results. However, due to the limited field-of-view of the pinhole camera, the performance quickly degrades when large camera movements are adopted. In this paper, we make the first attempt to generate novel views from a single indoor panorama and take the large camera translations into consideration. To tackle this challenging problem, we first use Convolutional Neural Networks (CNNs) to extract the deep features and estimate the depth map from the source-view image. Then, we leverage the room layout prior, a strong structural constraint of the indoor scene, to guide the generation of target views. More concretely, we estimate the room layout in the source view and transform it into the target viewpoint as guidance. Meanwhile, we also constrain the room layout of the generated target-view images to enforce geometric consistency. To validate the effectiveness of our method, we further build a large-scale photo-realistic dataset containing both small and large camera translations. The experimental results on our challenging dataset demonstrate that our method achieves state-of-the-art performance. The project page is at \url{https://github.com/bluestyle97/PNVS}.
\end{abstract}

\section{Introduction}

With the popularity of \SI{360}{\degree} cameras, panoramas have been widely used in many emerging domains such as Virtual Reality (VR). In a typical VR application, the device displays a \SI{360}{\degree} virtual scene, which can respond to 6 degree-of-freedom (DoF) head motion and give the user an immersive feeling. However, owing to the tedious image collection process, the panoramas are usually captured at a limited set of locations in practice, which restricts the DoF of scene viewing. With the expectation of providing a free-viewpoint scene visualization experience, we make the first attempt to address the problem of panoramic novel view synthesis from a single panorama.

\begin{figure}[t]
	\centering
	\includegraphics[width=\linewidth]{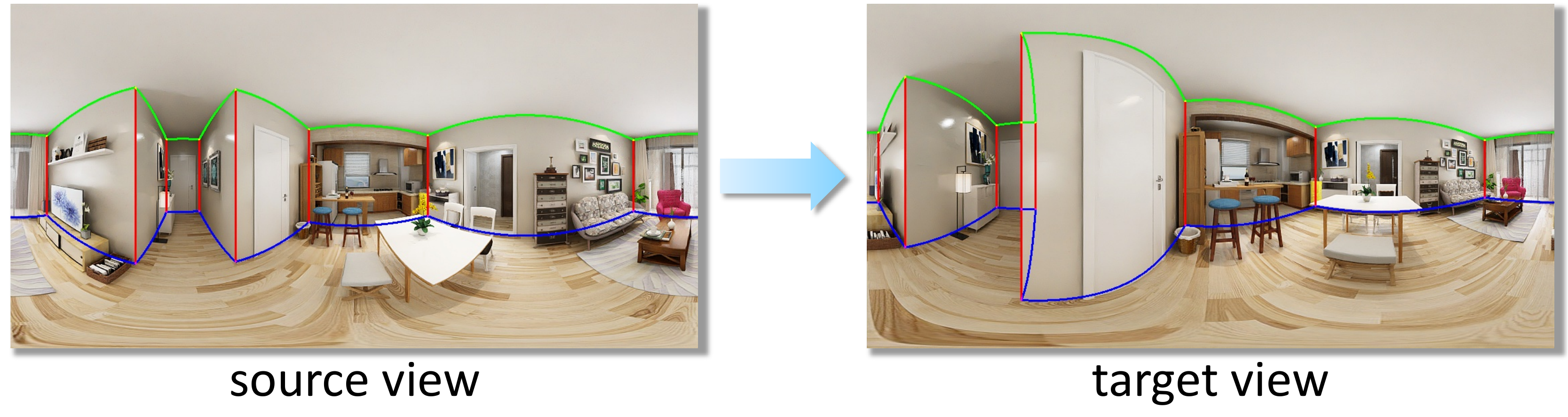}
	\includegraphics[width=\linewidth]{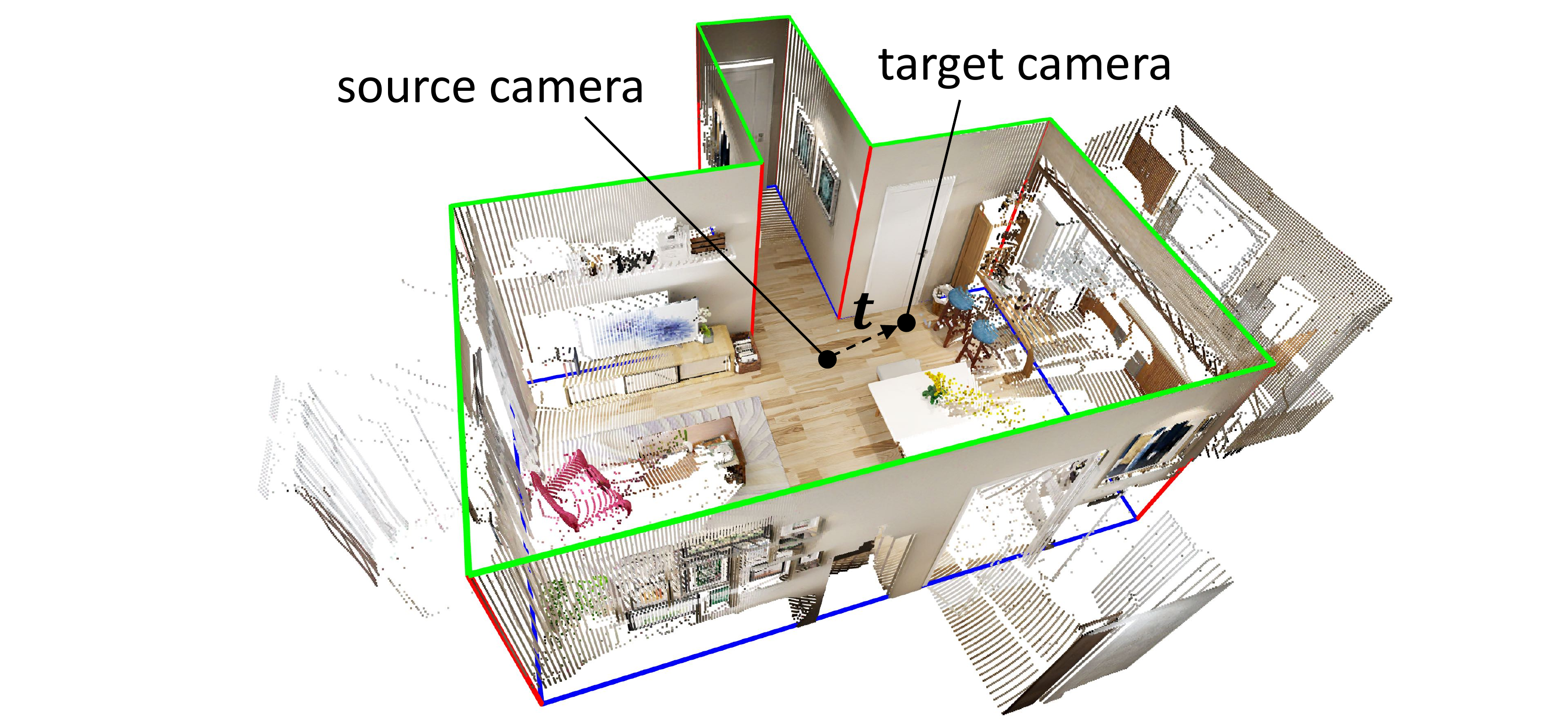}
	\caption{Panoramic novel view synthesis. Our goal is to generate a target-view panorama from the source-view panorama with camera translation $\vec{t}$. The green, red, and blue lines represent the ceiling-wall boundaries, wall-wall boundaries, and floor-wall boundaries of the room layout, respectively.}
	\label{fig:vis}
\end{figure}

In this paper, we constrain the panoramic view synthesis problem in the indoor scenario on account of its commonness in typical applications. Previous work~\cite{WilesGSJ20, TuckerS20, ShihSKH20} has shown promising results on novel view synthesis from a single perspective image. However, the performance quickly degrades when larger camera rotations and translations are adopted. Due to the limited field-of-view (FoV) of a pinhole camera, it is arduous to extrapolate the large unseen areas caused by violent camera motion. In contrast, a panorama inherently supports the rotational viewpoint change. Thus, we only need to consider camera translations. Furthermore, \SI{360}{\degree} FoV provides omnidirectional information, making it possible to consider larger camera translations. By synthesizing panoramic novel views, we can create new \SI{360}{\degree} contents to achieve 6-DoF scene viewing, which could potentially benefit many applications, such as virtual house tours.

The main challenge of novel view synthesis lies in recovering the missing areas caused by viewpoint change, and the difficulty is amplified when considering large camera translations. Fortunately, a panorama contains more structural information than a perspective image that can be exploited to reduce the difficulty. Previous work on image inpainting~\cite{NazeriNJQE19, RenYZLLL19} has proven the effectiveness of structural information to guide the content generation process. In the indoor scenario, the most common and easy-obtained structural information is the room layout, \ie, the ceiling-wall boundaries, floor-wall boundaries, and wall-wall boundaries. The synthesized images have to keep the room layout reasonable, especially when large camera translations are adopted.

Inspired by the state-of-the-art view synthesis framework~\cite{WilesGSJ20}, we propose a novel method to tackle the panoramic view synthesis problem and exploit the room layout as a prior and geometric constraint. The proposed method is composed of three stages. In the first stage, we use CNNs to extract a dense feature map, a depth map, and room layout from the source-view panorama. In the second stage, we transform the extracted feature map and room layout into the target view with a spherical geometric transformation process and fuse them to synthesize the target panorama. In the final stage, we estimate the room layout of the synthesized panorama and enforce the estimated layout consistent with the transformed target-view layout in the preceding stage.

To validate the effectiveness of our method and facilitate the research on this novel task, we further build a large-scale photo-realistic dataset upon Structured3D dataset~\cite{ZhengZLTGZ20}. The rendered images are high-fidelity, making the dataset close to realistic application scenarios. Besides the typical settings of previous work, our dataset also considers large camera translations to push the boundaries of the view synthesis task. We split our dataset into an easy set and a hard set according to the camera translation. The easy set contains target panoramas with small camera translations ranging from \SIrange{0.2}{0.3}{\meter}, including \SI{13080} training images and \SI{1791} testing images. The hard set contains target panoramas with large camera translations ranging from \SIrange{1}{2}{\meter}, including \SI{17661} training images and \SI{2279} testing images.

In summary, the main contributions of this paper are as follows: (i) We are the first to tackle the problem of synthesizing panoramic novel views from a single indoor panorama. (ii) We propose a novel layout-guided method to tackle this challenging task, which is able to handle large camera translations. (iii) We build a new high-quality and challenging dataset for this novel task, which contains small and large camera translations. (iv) The experimental results demonstrate that our method achieves state-of-the-art performance on this novel task and can be generalized to real datasets.

\section{Related Work}

\PARbegin{Novel view synthesis.} Previous work on novel view synthesis is based on heterogeneous settings, and we concentrate on learning-based methods here. The most straightforward idea is to perform image generation directly~\cite{HuangZLH2017, ZhaoWCLJF2018}. Instead, some methods~\cite{ZhouTSME16, ParkYYCB2017, ZhuSWCY2018} estimate the 2D correspondences between the source image and the target image first, \ie, appearance flows, to tackle this problem. More intuitively, many methods adopt the modeling-rendering pattern, which means modeling the scene first and then rendering it to novel views. Following this scheme, a variety of middle representations have been exploited, such as point cloud~\cite{NovotnyGR19, WilesGSJ20}, learned representations~\cite{PennerZ17, SitzmannTHNWZ2019, ChoiGTKK19}, layered depth image (LDI)~\cite{TulsianiTS18, ShihSKH20}, multi-plane images (MPI)~\cite{ZhouTFFS2018, FlynnBDDFOST2019, MildenhallSOKRNK2019, SrinivasanTBRNS2019, TuckerS20, LiuZMR20} and neural radiance fields~\cite{MildenhallSTBRN20, ZhangRSK2020}.

Compared with common perspective settings, attempts on view synthesis from panoramas are still very limited so far. Some previous work~\cite{HuangCCJ17, SerranoKCDGHM19} has tackled the problem of 6-DoF viewing from a pre-captured \SI{360}{\degree} video to promote VR applications. \citet{HuangCCJ17} propose to reconstruct a point cloud from the input \SI{360}{\degree} video to achieve real-time 6-DoF video playback with a VR device. \citet{SerranoKCDGHM19} present a method for adding parallax and real-time playback of \SI{360}{\degree} videos, which relies on a layered scene representation. Recently, inspired by the MPI representation, \citet{LinXMSHDSSR20} and \citet{AttalLGRT20} propose multi-depth panorama (MDP) and multi-sphere image (MSI) representation respectively, to conduct 6-DoF rendering from \SI{360}{\degree} imagery. However, their settings are quite different from ours. \citet{LinXMSHDSSR20} take the images captured by a multi-camera \SI{360}{\degree} panorama capture rigs as input, while the input of~\cite{AttalLGRT20} is a \SI{360}{\degree} stereo video.

\begin{figure*}[t]
	\centering
	\includegraphics[width=\linewidth]{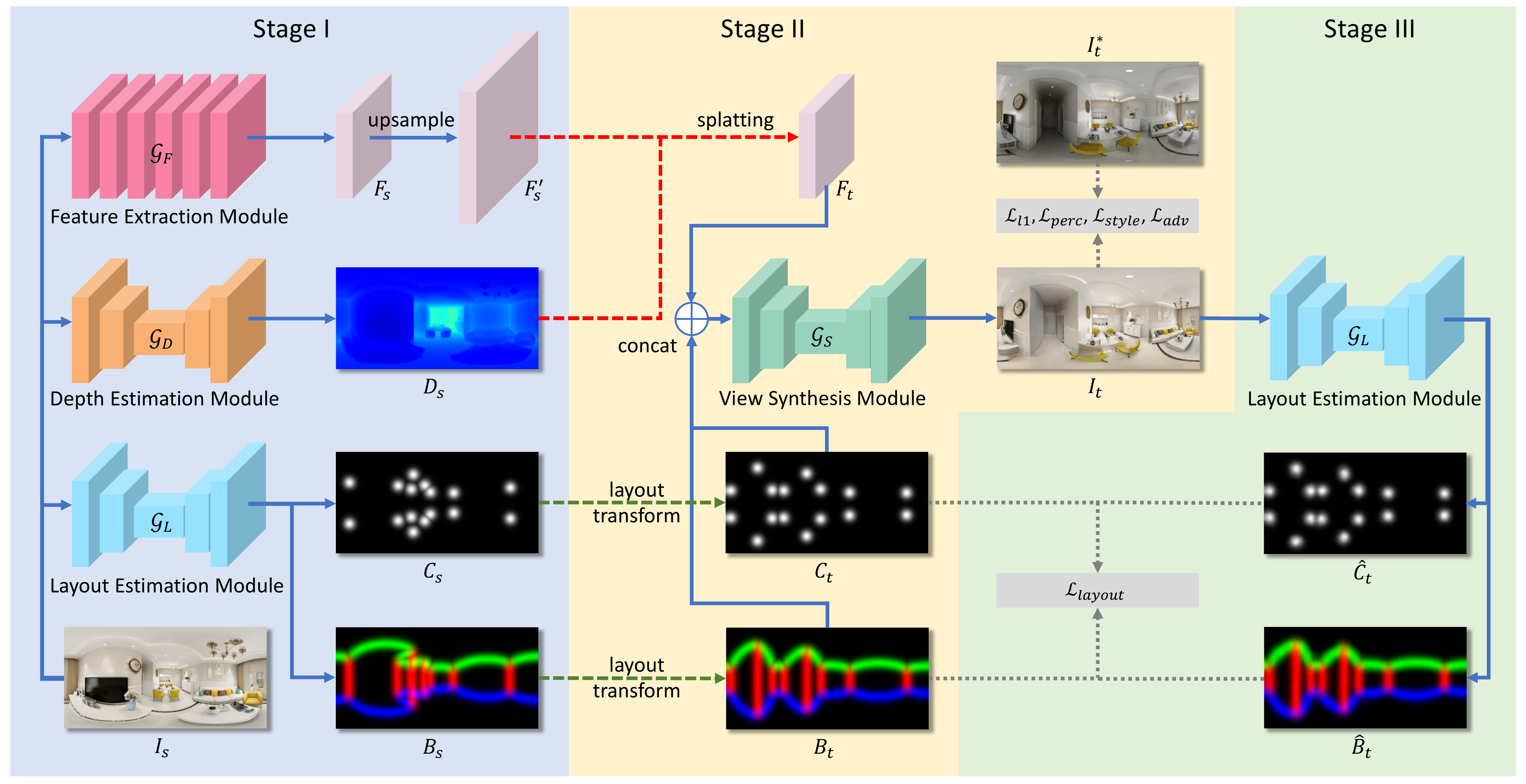}
	\caption{An overview of our pipeline. In the first stage, the network extracts a dense feature map $\mat{F}_s$ from the source-view panorama $\mat{I}_s$ as contextual information, and estimates its depth $\mat{D}_s$ as well as room layout $\mat{L}_s=\{\mat{B}_s, \mat{C}_s\}$ as structural information. In the second stage, $\mat{F}_s$ and $\mat{L}_s$ are transformed into the target viewpoint with a forward splatting operation and a layout transformation process to form $\mat{F}_t$ and $\mat{L}_t=\{\mat{B}_t, \mat{C}_t\}$, respectively. Then, $\mat{F}_t$ and $\mat{L}_t$ are fused together to synthesize the target-view panorama $\mat{I}_t$. In the final stage, we estimate the room layout of synthesized panorama $\mat{I}_t$ and enforce it consistent with the transformed layout $\mat{L}_t$.}
	\label{fig:pipeline}
\end{figure*}

\PAR{Image inpainting.} Image inpainting aims to complete the missing region in an image. Traditional patch-based methods~\cite{LiuYFG2018} and diffusion-based methods~\cite{SrideviK2019} are the pioneering work to tackle this problem. In the deep learning era, CNN-based methods~\cite{YanLLZS2018, LiuJXY2019, ZengFCG2019, LiWZDT2020} and GAN-based methods~\cite{YuLYSLH2018, YuLYSLH2019, LahiriJAMB2020, YiTAJX2020} draw more attention from the research community due to their favorable performance. Several inpainting methods have demonstrated the effectiveness of using structural information. Back in the non-deep-learning era, \citet{SunYJS2005} and \citet{HuangKAK2014} have proposed to use line and planar structure to guide the image inpainting process. Various learning-based methods~\cite{NazeriNJQE19, RenYZLLL19, XiongYLYLBL2019, LiHZDT2019} also exploit the structural information. Although the structural information differs in specific forms, \eg, edges, gradients, sketches, or foreground contours, they all act as a global structural prior as well as a geometric constraint and have shown reliable effectiveness.

\PAR{Layout and depth estimation on panoramas.} Room layout estimation from a panorama has been sufficiently studied. LayoutNet~\cite{ZouCSH18} predicts a boundary probability map and a corner probability map from the input panorama, then estimates the room layout with a Manhattan layout optimizer. HorizonNet~\cite{SunHSC19} further simplifies the layout representation by replacing the 2D probability maps with 1D vectors. DuLa-Net~\cite{YangWPWSC19} exploits an equirectangular panorama branch and a perspective ceiling-view branch to tackle this problem.

OmniDepth~\cite{ZioulisKZD18} transfers the monocular depth estimation task to panoramas first. \citet{ZioulisKZAD19} propose a self-supervised method to estimate panoramic depth, which uses panoramic view synthesis as a proxy task. BiFuse~\cite{WangYSCT20} adopts a two-branch architecture to predict panoramic depth. \citet{JinXZZTXYG20} propose to leverage the geometric structure of a scene, \ie, different room layout representations, to conduct this problem. Recently, \citet{ZengKG20} propose to jointly learn the panoramic layout and depth since they are tightly intertwined.

\section{Method}

Given a source-view panorama $\mat{I}_s \in \mathbb{R}^{H \times W \times 3}$ at the source camera position $\vec{p}_{s} \in \mathbb{R}^{3}$ and a target camera position $\vec{p}_t \in \mathbb{R}^{3}$, our goal is to synthesize a target-view panorama $\mat{I}_t \in \mathbb{R}^{H \times W \times 3}$. Since the panorama inherently support camera rotations, we can assume that the cameras always face the same direction and only consider the camera translations here.

Our method follows the classical modeling-rendering pattern. We first conduct depth estimation on the source-view image to obtain the 3D scene. Since the estimated 3D scene is inaccurate and noisy, directly rendering new views from it leads to severe shape distortion and pixel misalignment. Inspired by the recent success in room layout estimation~\cite{ZouCSH18, SunHSC19}, we exploit it as a structural prior and geometric constraint to guide the view synthesis process. The three-stage pipeline is shown in \figref{fig:pipeline}.

\subsection{Feature Extraction and Structure Estimation}

In the first stage, we extract contextual and structural information from the source-view image. Concretely, the feature extraction module $\mathcal{G}_F$ extracts a dense feature map $\mat{F}_s$ from $\mat{I}_s$ as contextual information, and the layout estimation module $\mathcal{G}_L$ estimates the room layout $\mat{L}_s$ from $\mat{I}_s$ as structural information. To build the scene geometry, the depth estimation module $\mathcal{G}_D$ predicts a depth map $\mat{D}_s$ from $\mat{I}_s$.

Previous work~\cite{WilesGSJ20} has shown that synthesizing novel views from high-level features containing scene semantics instead of simple RGB colors leads to better results. Following this spirit, our model utilizes a CNN $\mathcal{G}_F$ to extract a dense feature map $\mat{F}_s \in \mathbb{R}^{H \times W \times C}$ from the input RGB panorama $\mat{I}_s \in \mathbb{R}^{H \times W \times 3}$.

Similar to LayoutNet~\cite{ZouCSH18}, the layout estimation module $\mathcal{G}_L$ predicts a boundary map $\mat{B}_s \in \mathbb{R}^{H \times W \times 3}$ and a corner map $\mat{C}_s \in \mathbb{R}^{H \times W}$. With $\mat{B}_s$ and $\mat{C}_s$, we follow the standard post-processing procedure of LayoutNet to obtain the 2D positions of room corners $\mat{L}_s \in \mathbb{R}^{N \times 2}$.

The feature extraction, depth and layout estimation process can be represented as:
\begin{align}
	\mat{F}_s = \mathcal{G}_{F} (\mat{I}_s), \mat{D}_s = \mathcal{G}_{D} (\mat{I}_s), \{\mat{B}_s, \mat{C}_s\} = \mathcal{G}_{L} (\mat{I}_s).
\end{align}

$\mathcal{G}_F$ is implemented as a series of ResNet blocks, and $C$ is set to \num{64}. We follow the architectures of \citet{HuOZO19} and LayoutNet~\cite{ZouCSH18} to implement $\mathcal{G}_D$ and $\mathcal{G}_L$, respectively.

\begin{figure}[t]
	\centering
	\includegraphics[width=\linewidth]{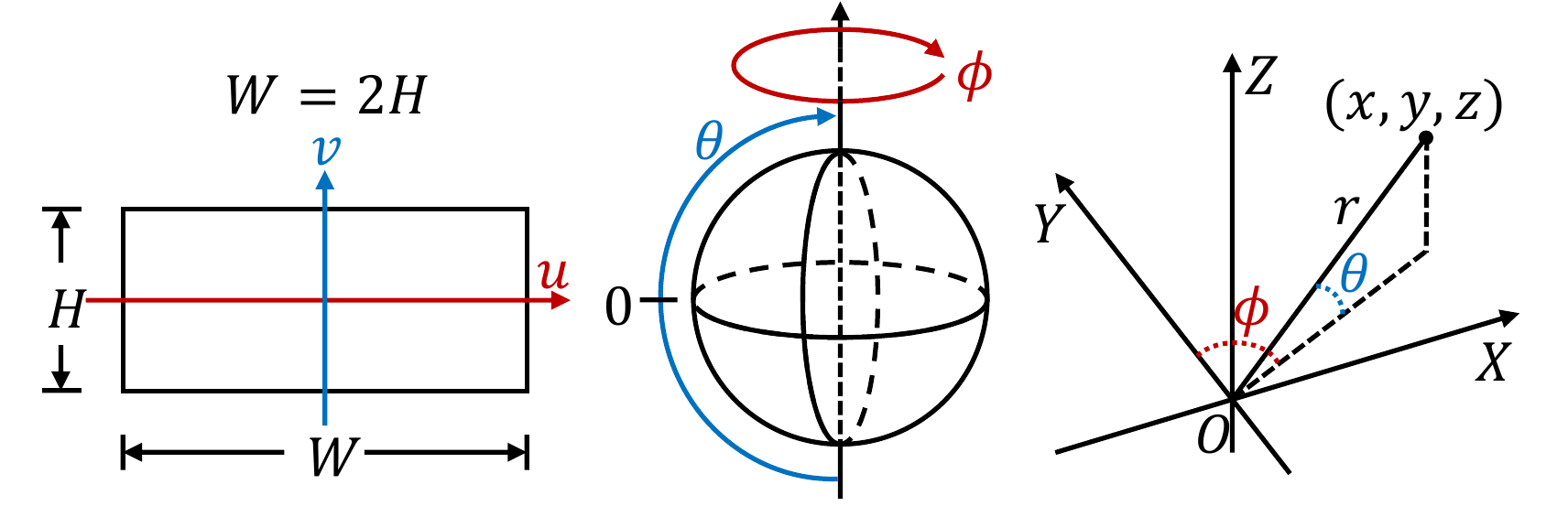}
	\caption{The geometric interpretation of the relationships among $\mathcal{P}$, $\mathcal{S}$, and $\mathcal{C}$. The left and middle pictures explain the relationship between coordinates $(u, v) \in \mathcal{P}$ and $(\phi, \theta) \in \mathcal{S}$, which are borrowed from \cite{Xiao2012}. The right picture explains the relationship between coordinates $(\phi, \theta, r) \in \mathcal{S}$ and $(x, y, z) \in \mathcal{C}$, $O$ is the camera center.}
	\label{fig:coord}
\end{figure}

\subsection{Viewpoint Transformation and View Synthesis with Layout Prior}

In the second stage, we transform the source-view contextual $\mat{F}_s$ and structural information $\mat{L}_s$ into the target view and synthesize the target panorama $\mat{I}_t$.

The viewpoint transformation is a spherical geometric transformation process. To make it easier to understand, we first clarify several related coordinate systems and show their relationships in \figref{fig:coord}.

\begin{itemize}
	\item {\bf Panoramic pixel grid coordinate system} $\mathcal{P}$: Coordinates $(u, v) \in \mathcal{P}$ represent the pixel at the $u$-th column and the $v$-th row pixel on the panoramic image plane, where $u \in [0, W)$ and $v \in [0, H)$.

	\item {\bf Spherical polar coordinate system} $\mathcal{S}$: The origin is the camera position. Coordinates $(\phi, \theta, r) \in \mathcal{S}$ represents a point whose longitude is $\phi$, latitude is $\theta$, and distance from the origin is $r$, where $\phi \in [-\pi,\pi]$, $\theta \in [-\pi/2, \pi/2]$, $r>0$.

	\item {\bf 3D Cartesian camera coordinate system} $\mathcal{C}$: The origin is the camera position. The $X, Y, Z$ axes points rightward, forward and upward, respectively. Coordinates $(x,y,z) \in \mathcal{C}$ represents the position of a 3D point relative to the origin, where $x,y,z \in \mathbb{R}$.
\end{itemize}

\PAR{Feature map view transformation.} To transform the source-view feature map $\mat{F}_s$ into the target view, we need to map each source-view pixel $(u_s, v_s) \in \mathcal{P}_{s}$ to a target-view pixel $(u_t, v_t) \in \mathcal{P}_{t}$, which can be accomplished by a series of coordinate transformations from $\mathcal{P}_{s}$ to $\mathcal{P}_{t}$:
\begin{align}
	f = f_{\mathcal{S}_t \mapsto \mathcal{P}_t} \circ f_{\mathcal{C}_t \mapsto \mathcal{S}_t} \circ f_{\mathcal{C}_s \mapsto \mathcal{C}_t} \circ f_{\mathcal{S}_s \mapsto \mathcal{C}_s} \circ f_{\mathcal{P}_{s} \mapsto \mathcal{S}_s},
	\label{eq:correspondence:expand}
\end{align}
where $f_{A\mapsto B}$ denotes a coordinate transformation from coordinate system $A$ to $B$, and $\circ$ denotes the composition of transformations. We refer the readers to supplementary material for detailed coordinate transformation equations.

By conducting \eqref{eq:correspondence:expand} on a source-view pixel $(u_s,v_s) \in \mathcal{P}_s$, we can obtain its corresponding target-view pixel position $(u_t,v_t) = f\left(u_s,v_s\right) \in \mathcal{P}_t$. With the pixel correspondences, we adopt a differentiable rendering approach~\cite{TulsianiTS18, ZioulisKZAD19} to generate target-view feature map $\mat{F}_t \in \mathbb{R}^{H\times W\times C}$. Concretely, we splat the feature vector at each pixel of $\mat{F}_s$ onto its corresponding pixel position on the target-view panorama plane with the bilinear interpolation. To resolve the conflicts caused by the many-to-one mapping problem, a soft z-buffering is adopted, which can be formulated as:
\begin{align}
	\mat{I}_t(u_t, v_t) = \frac{ \sum_{(u_s, v_s)} \mat{I}_s (u_s, v_s) \exp (-\mat{D}_s(u_s, v_s) / d_{\max })}{ \sum_{(u_s, v_s)} \exp ( - \mat{D}_s (u_s, v_s) / d_{\max} ) + \epsilon},
	\label{eq:render}
\end{align}
where $d_{\max} = 10$ is a pre-defined maximum depth value, $\epsilon$ is a small constant for numerical stability.

\begin{figure}[t]
	\centering
	\includegraphics[width=\linewidth]{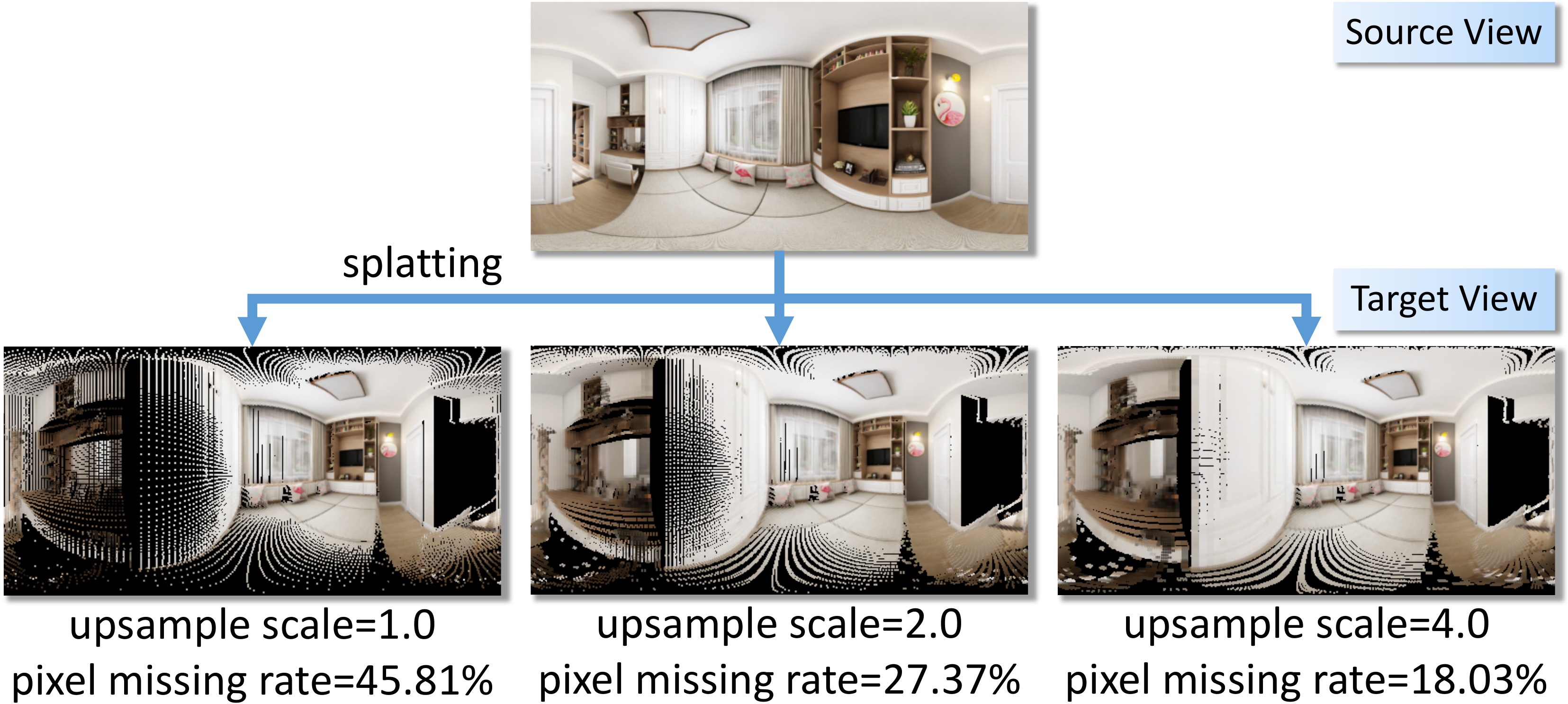}
	\caption{The influence of feature map upsampling. In the splatting operation, each source-view pixel contributes to 4 neighboring target pixels. Therefore, using a denser source-view feature map leads to smaller holes on the splatted feature map. Since it is hard to visualize the feature, we show the splatting results of RGB values under different upsampling scales instead.}
	\label{fig:upsample}
\end{figure}

\PAR{Feature map upsampling.} Since large camera translations are taken into consideration, directly splatting leads to large missing areas (\ie, holes) on $\mat{F}_t$, making it difficult to inpaint. Besides, some areas that are supposed to be occluded will be unexpectedly exposed because the areas occluding them are missing. To tackle this problem, we upsample $\mat{F}_s$ to $\mat{F}_s'$ before the forward splatting:
\begin{align}
	\mat{F}_s' = \operatorname{Conv} ( \operatorname{Upsample} ( \mat{F}_s )),
	\label{eq:upsample}
\end{align}
where $\operatorname{Conv}$ denotes a convolution layer, $\operatorname{Upsample}: \mathbb{R}^{H \times W \times C} \rightarrow \mathbb{R}^{2H \times 2W \times C}$  denotes the nearest upsampling layer. This operation can significantly reduce the missing areas in $\mat{F}_t$ and make it easier for the network to inpaint. \figref{fig:upsample} demonstrates such effect.

\PAR{Layout view transformation.} The layout transformation from $\vec{L}_s \in \mathbb{R}^{N\times 2}$ to $\vec{L}_t \in \mathbb{R}^{N\times 2}$ is similar to feature map transformation but have some differences. Be aware that we cannot know the depths of layout corners from $\mat{D}_s$ since they may be occluded by foreground objects. Thus, we estimate the depth of each corner with the camera height $h$. We provide the details of the layout transformation process in the supplementary material.

To utilize $\mat{L}_t$, we draw a boundary map $\mat{B}_t \in \mathbb{R}^{H \times W \times 3}$ and a corner map $\mat{C}_t \in \mathbb{R}^{H \times W}$ from $\mat{L}_t$ with Gaussian blurring. Then, we feed them into the view synthesis module $\mathcal{G}_S$ to serve as structural prior and constrain the synthesis of target-view panorama.

\PAR{View synthesis with layout prior.} With the transformed target-view contextual information $\mat{F}_t$ and structural information $\{\mat{B}_t, \mat{C}_t\}$, the view synthesis module $\mathcal{G}_S$ fuses them all together and synthesizes the target-view panorama $\mat{I}_t$:
\begin{align}
	\mat{I}_t = \mathcal{G}_{S} (\mat{F}_t \oplus \mat{B}_t \oplus \mat{C}_t),
\end{align}
where $\oplus$ denotes the concatenation operation along the channel dimension. We adopt an architecture similar to \cite{NazeriNJQE19} to implement $\mathcal{G}_S$.

\subsection{Layout Consistency Constraint}

In order to maximize the use of room layout guidance, we introduce a layout-consistency loss to force the synthesized panorama $\mat{I}_t$ to keep the consistency of room layout. Specifically, we feed $\mat{I}_t$ into the layout estimation module to obtain $\hat{\mat{B}_t}$ and $\hat{\mat{C}_t}$. Then, we compare them with $\mat{B}_t, \mat{C}_t$ and calculate the layout consistency loss as:
\begin{align}
	\mathcal{L}_\textrm{layout} = \operatorname{BCE} (\hat{\mat{B}_t}, \mat{B}_{t} ) + \operatorname{BCE} (\hat{\mat{C}_t}, \mat{C}_{t} ),
\end{align}
where $\operatorname{BCE}$ represents the binary cross entropy loss.

\subsection{Losses}

During training, the layout estimation module $\mathcal{G}_L$ and the depth estimation module $\mathcal{G}_D$ are pretrained under the supervision of ground-truth layout and depth, respectively.

Given the synthesized panorama $\mat{I}_{t}$ and the ground-truth panorama $\mat{I}_{t}^{*}$, the rest model is trained with $\ell_1$ loss, perceptual loss~\cite{JohnsonAF16}, style loss~\cite{GatysEB16}, adversarial loss~\cite{Goodfellow2014} and layout consistency loss. Their functions can be formulated as:
\begin{align}
	\mathcal{L}_{\ell_1} & =\mathbb{E}\left[\left\|\boldsymbol{I}_{t}-\boldsymbol{I}_{t}^{*}\right\|_{1}\right],\\
	\mathcal{L}_{\text {perc}} & =\mathbb{E}\left[\sum_{i} \left\|\psi_{i}\left(\boldsymbol{I}_{t}\right)-\psi_{i}\left(\boldsymbol{I}_{t}^{*}\right)\right\|_{1}\right],\\
	\mathcal{L}_{\text {style }} & =\mathbb{E}\left[\sum_{j}\left\|G_{j}\left(\boldsymbol{I}_{t}\right)-G_{j}\left(\boldsymbol{I}_{t}^{*}\right)\right\|_{1}\right],\\
	\mathcal{L}_{\text{adv}} & =\mathbb{E}\left[\log \mathcal{D}\left(\boldsymbol{I}_{t}^{*}\right)\right]+\mathbb{E}\left[\log \left(1-\mathcal{D}\left(\mathcal{G}\left(\boldsymbol{I}_{s}\right)\right)\right)\right],
\end{align}
where $\psi_{i}$ denotes the activation map of the $i$-th layer of a pretrained VGG-19, $G_{j}$ is a $C_{j} \times C_{j}$ Gram matrix calculated from $\psi_{j}$. $\mathcal{G}$ denotes the generator, i.e., our model, and $\mathcal{D}$ denotes the discriminator, $\boldsymbol{I}_{t}=\mathcal{G}\left(\boldsymbol{I}_{s}\right)$.

Finally, the total loss is calculated as:
\begin{align}
	\mathcal{L} = \mathcal{L}_{\ell_1} + \mathcal{L}_\textrm{perc} + \lambda \mathcal{L}_\textrm{style} + \mathcal{L}_\textrm{adv} + \mathcal{L}_\textrm{layout},
\end{align}
where $\lambda$ is set to \num{100} in our experiments.

\begin{figure*}[t]
	\centering
	\includegraphics[width=\linewidth]{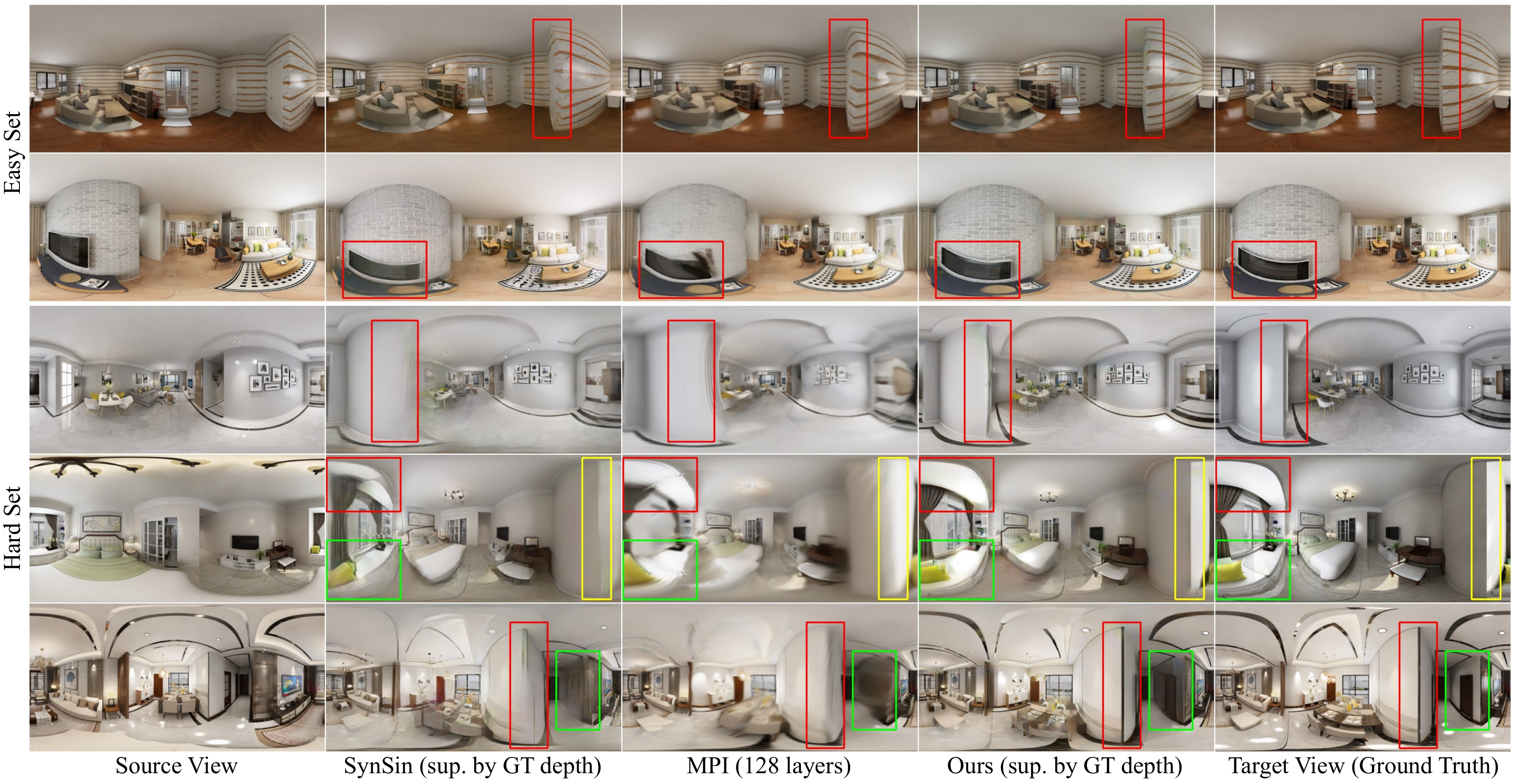}
	\caption{Qualitative view synthesis results on our dataset. The first two rows are from the easy set, while the last three rows are from the hard set. We highlight the major differences using the bounding boxes. More results are shown in the supplementary material.}
	\label{fig:results}
\end{figure*}

\begin{table*}[t]
	\centering
	\sisetup{
		table-number-alignment = center,
		table-auto-round = true,
		table-figures-decimal = 4,
		detect-weight = true,
		detect-inline-weight = math
	}
	\begin{tabular}{lS[table-figures-decimal=2]SSS[table-figures-decimal=2]SS}
		\toprule
		\multirow{2}{*}{Methods} & \multicolumn{3}{c}{Easy Set (\SIrange{0.2}{0.3}{\meter})} & \multicolumn{3}{c}{Hard Set (\SIrange{1.0}{2.0}{\meter})} \tabularnewline
		\cmidrule(lr){2-4} \cmidrule(lr){5-7}
		& {PSNR $\uparrow$} & {SSIM $\uparrow$} & {LPIPS $\downarrow$} & {PSNR $\uparrow$} & {SSIM $\uparrow$} & {LPIPS $\downarrow$} \tabularnewline
		\midrule
		{SynSin (end-to-end)} & 16.875742 & 0.743338 & 0.194632 & 15.512907 & 0.729772 & 0.246206 \tabularnewline
		{SynSin (supervised by GT depth)} & 18.035078 & 0.785327 & 0.171387 & 17.022964 & 0.782655 & 0.211907 \tabularnewline
		{SynSin (GT depth as input)} & 18.794199 & 0.812714 & 0.15589 & 18.019827 & 0.818087 & 0.172375 \tabularnewline
		\midrule
		{MPI (32 layers)} & 18.321579 & 0.804379 & 0.214988 & 16.530794 & 0.772471 & 0.309808 \tabularnewline
		{MPI (64 layers)} & 18.077526 & 0.798378 & 0.219242 & 16.561275 & 0.776904 & 0.305071 \tabularnewline
		{MPI (128 layers)} & 18.230545 & 0.801472 & 0.217019 & 16.502985 & 0.777618 & 0.301465 \tabularnewline
		\midrule
		{Ours (supervised by GT depth)} & 19.349605 & 0.837295 & 0.135129 & 17.495288 & 0.814818 & 0.176931 \tabularnewline
		{Ours (GT depth as input)} & 20.52127 & 0.872651 & 0.119191 & 18.533648 & 0.855205 & 0.154397 \tabularnewline
		{Ours (GT depth \& GT layout as input)} & 20.825109 & 0.874334 & 0.114998 & 18.947082 & 0.859285 & 0.145394 \tabularnewline
		\bottomrule
	\end{tabular}
	\caption{Quantitative results on our dataset.}
	\label{tab:results}
\end{table*}

\section{Experiments}

In this section, we conduct experiments to validate the performance of our proposed method. Due to the space limitation, we refer the readers to the supplementary material for extensive qualitative results and failure cases.

\begin{figure*}[t]
	\centering
	\includegraphics[width=\linewidth]{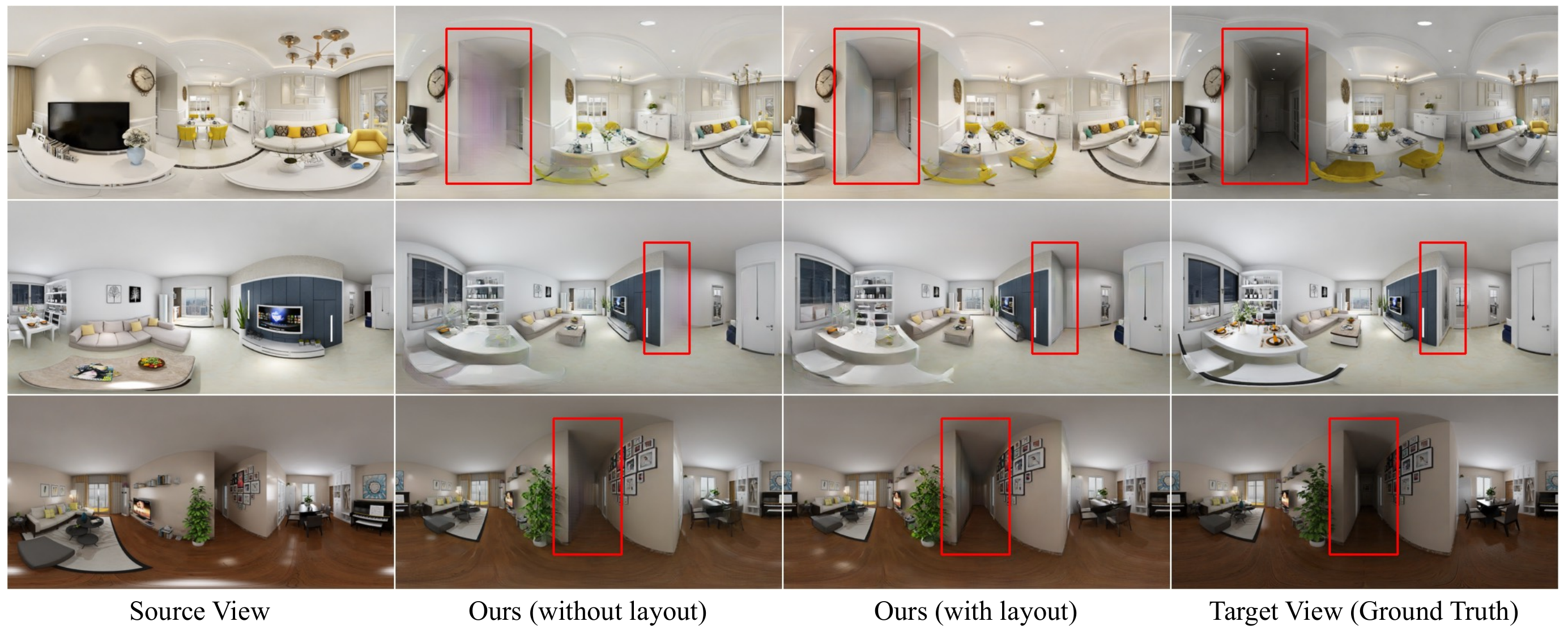}
	\caption{The effectiveness of room layout guidance. We highlight the major differences using the bounding boxes.}
	\label{fig:layout}
\end{figure*}

\subsection{Experimental Setup}

\PARbegin{Implementation details.} Our model is implemented with PyTorch library~\cite{PyTorch} and trained on two NVIDIA TITAN V GPU devices. We use the Adam~\cite{KingmaB15} optimizer with $\beta_1 = 0.9$ and $\beta_2 = 0.999$. The batch size is set to \num{4}. Specifically, we first train the depth estimation module and the layout estimation module for \num{30} epochs to make them converge. Then, we freeze them and train the rest model for another \num{50} epochs. The learning rate for both the generator and the discriminator is set to \num{1e-4}. After \num{30} epochs, we reduce the learning rate by \num{10} times.

\PAR{Dataset.} Our panoramic view synthesis dataset is built upon Structured3D dataset~\cite{ZhengZLTGZ20}. Each panorama in Structured3D corresponds to a different room. We regard original images as source views and render three target views for each source view. Our dataset is divided into two sets with different target-view camera selection strategies: (i) \textbf{an easy set}: the camera translation ranges from \SIrange{0.2}{0.3}{\meter} along random directions, which is a typical translational distance in previous view synthesis work. (ii) \textbf{a hard set}: the camera translation ranges from \SIrange{1.0}{2.0}{\meter} along random directions, which is a very challenging setting and has rarely been considered. To clarify the difficulties of our settings, we visualize the relationship between the pixel missing rate after the splatting operation and the camera translation distance in the supplementary material. The resolution of the panorama in our dataset is $512 \times 1024$. In all experiments, we take panoramas of $256 \times 512$ as input.

\PAR{Evaluation metrics.} We quantify the performance of our method with three metrics: (i) Peek Signal-to-Noise Ratio (PSNR), (ii) Structural Similarity (SSIM), and (iii) Learned Perceptual Image Patch Similarity (LPIPS)~\cite{ZhangIESW18}.

\subsection{Experimental Results}

\PARbegin{Methods for comparison.} We compare our approach with two state-of-the-art single-image view synthesis methods: point-cloud-based method SynSin~\cite{WilesGSJ20}\footnote{\url{https://github.com/facebookresearch/synsin}} and MPI-based method~\cite{TuckerS20}\footnote{\url{https://github.com/google-research/google-research/tree/master/single_view_mpi}} (in short, MPI). We choose SynSin~\cite{WilesGSJ20} and MPI~\cite{TuckerS20} as our baselines on account of their good performance as well as the code availability and portability.

We modify SynSin and MPI to make them applicable to panoramas. For SynSin, the perspective projection in the differentiable renderer is replaced with equirectangular projection. Every 3D point is projected to a circular region of the target-view panorama plane with $\alpha$-compositing. For MPI, we use the same network as~\cite{TuckerS20} to infer a multi-sphere image (MSI) centered at the camera position, which is similar to~\cite{AttalLGRT20}. Then, we cast the rays from the target view onto the MSI and use the bilinear interpolation to perform view synthesis.

\PAR{Quantitative evaluation.} SynSin estimates the depth of the source view with an end-to-end training scheme. For a fair comparison, we also train SynSin with ground-truth depth as supervision to meet our setting. Besides, we evaluate SynSin and our model with ground-truth depth as input to investigate their upper-bound performance. For MPI, we set the number of layers as \SI{32}, \SI{64}, and \SI{128}, respectively.

As \tabref{tab:results} shows, the performance of SynSin and our model increases when using more accurate depth, and the performance of MPI increases when more layers are adopted. Our method outperforms the other two methods in all metrics. When exploiting ground-truth depth as input, our model shows a higher upper bound than SynSin. In addition, we can further boost the performance of our model by adopting the ground-truth layout as input.

\PAR{Qualitative evaluation.} \figref{fig:results} shows the qualitative results of three methods. As one can see, our approach maintains more plausible visual details. Especially the results on the hard set show that our method can maintain the room structure well when large camera translation is adopted. However, the other two approaches have artifacts, such as blurring layout boundaries and distortion.

\PAR{User study.} For more complete qualitative comparison, we further conduct a user study. We first sample \num{50} images from the easy set and the hard set, respectively. Then, we recruit \num{50} volunteers and show them the synthesized target views of the three methods in random order, and the ground truth. We ask them to select the closest one to the ground truth and report the percentage of volunteers who prefer a given method. As shown in \tabref{tab:user}, volunteers prefer our method over the other two methods in both sets.

\begin{table}[t]
	\centering
	\sisetup{
		table-number-alignment = center,
		table-auto-round = true,
		table-figures-decimal = 2,
		detect-weight = true,
		detect-inline-weight = math
	}
	\begin{tabular}{lSS}
		\toprule
		\multirow{2}{*}{Methods} & {Easy Set} & {Hard Set} \tabularnewline
		& {(\SIrange{0.2}{0.3}{\meter})} & {(\SIrange{1.0}{2.0}{\meter})} \tabularnewline
		\midrule
		{SynSin} & 0.21 & 0.18 \tabularnewline
		{MPI} & 0.13 & 0.09 \tabularnewline
		{Ours} & \bfseries 0.66 & \bfseries 0.73 \tabularnewline
		\bottomrule
	\end{tabular}
	\caption{User study on our dataset.}
	\label{tab:user}
\end{table}

\subsection{Ablation Studies}

\begin{figure*}[t]
	\centering
	\includegraphics[width=\linewidth]{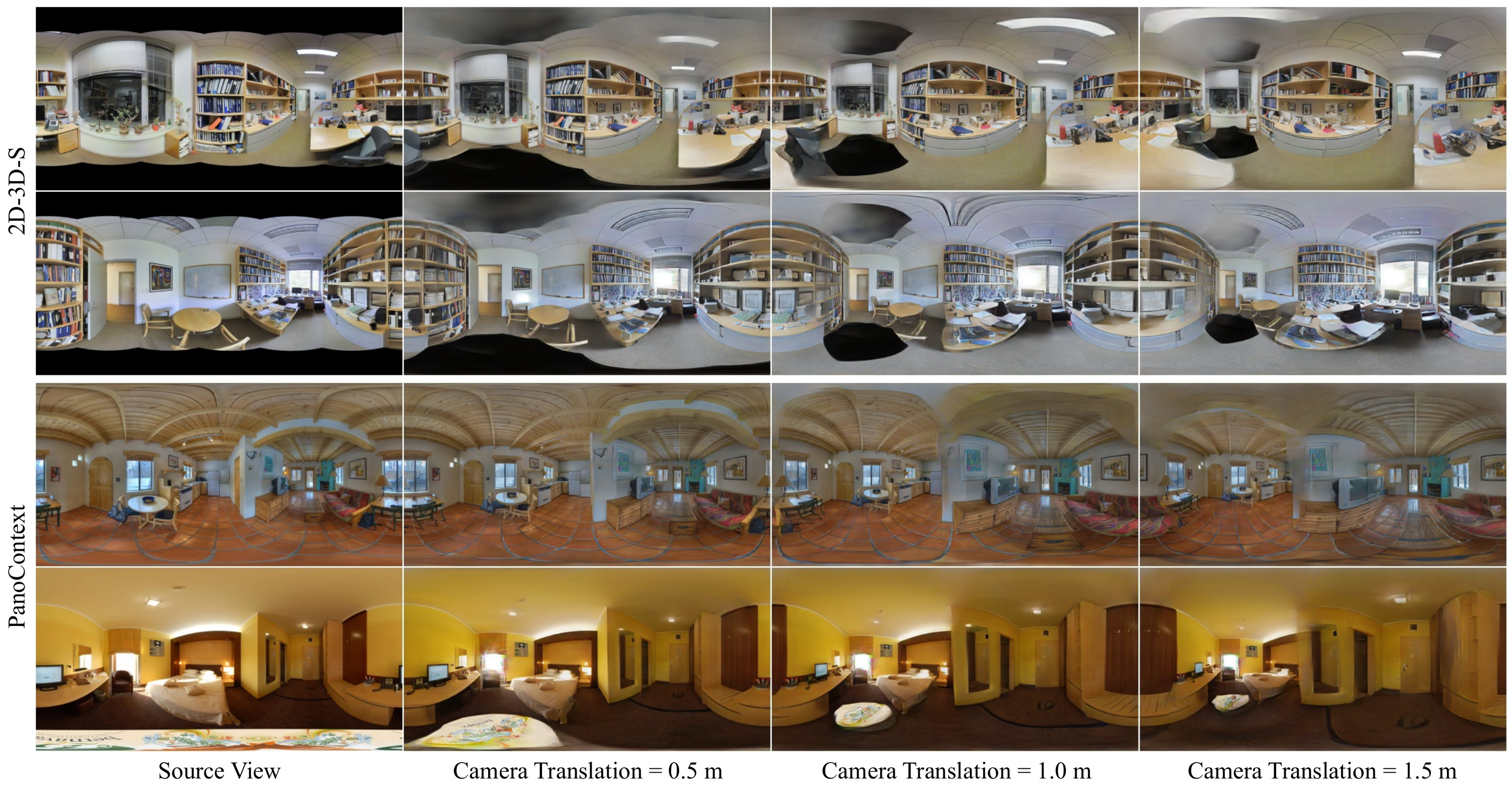}
	\caption{Panoramic view synthesis on 2D-3D-S dataset and PanoContext dataset. More results are shown in the supplementary material.}
	\label{fig:inference}
\end{figure*}

\begin{table*}[t]
	\centering
	\sisetup{
		table-number-alignment = center,
		table-auto-round = true,
		table-figures-decimal = 4,
		detect-weight = true,
		detect-inline-weight = math
	}
	\begin{tabular}{cccS[table-figures-decimal=2]SSS[table-figures-decimal=2]SS}
		\toprule
		\multicolumn{3}{c}{Components} & \multicolumn{3}{c}{Easy Set (\SIrange{0.2}{0.3}{\meter})} & \multicolumn{3}{c}{Hard Set (\SIrange{1.0}{2.0}{\meter})} \tabularnewline
		\cmidrule(lr){1-3} \cmidrule(lr){4-6} \cmidrule(lr){7-9}
		{Upsampling} & {Prior} & {Consistency} & {PSNR $\uparrow$} & {SSIM $\uparrow$} & {LPIPS $\downarrow$} & {PSNR $\uparrow$} & {SSIM $\uparrow$} & {LPIPS $\downarrow$} \tabularnewline
		\midrule
		& & & 18.443278 & 0.806734 & 0.151122 & 16.477028 & 0.773478 & 0.200652 \tabularnewline
		$\checkmark$ & & & 18.571305 & 0.810991 & 0.15043 & 16.69582 & 0.780184 & 0.196523 \tabularnewline
		$\checkmark$ & & $\checkmark$ & 18.492516 & 0.81725 & 0.150275 & 16.934402 & 0.798107 & 0.206624 \tabularnewline
		$\checkmark$ & $\checkmark$ & & \bfseries 19.39685 & 0.837107 & 0.145468 & 17.461016 &0.814602 & 0.191674 \tabularnewline
		& $\checkmark$ & $\checkmark$ & 19.115038 & 0.835439 & 0.139029 & 17.326105 & 0.808974 & 0.18125 \tabularnewline
		$\checkmark$ & $\checkmark$ & $\checkmark$ & 19.349605 & \bfseries 0.837295 & \bfseries 0.135129 & \bfseries 17.495288 & \bfseries 0.814818 & \bfseries 0.176931 \tabularnewline
		\bottomrule
	\end{tabular}
	\caption{Ablation studies on our dataset.}
	\label{tab:ablation}
\end{table*}

We conduct some ablation studies to verify the effectiveness of each component in our proposed method. The results are shown in \tabref{tab:ablation} and discussed in detail next.

\PAR{Feature map upsampling.} We try to remove the feature map upsampling from our model. By comparing the quantitative results with the complete model, we can see that the upsampling operation leads to a performance improvement. As shown in \figref{fig:upsample}, the upsampling operation can remarkably reduce the missing pixels after the splatting operation, which abates the contextual information loss and makes the inpainting easier for the view synthesis module.

\PAR{Room layout.} To show the effectiveness of room layout, we remove either the layout prior or layout consistency loss, or both of them. The results show that both the layout prior and layout consistency loss contribute to the performance improvement. When using the layout prior, all metrics increase by a large margin. The layout consistency loss leads to better perceptual quality, which is indicated by the improvement of LPIPS. When using both of them, the performance of the model reaches the peak. Besides, \figref{fig:layout} visualizes some target-view results synthesized with or without layout guidance on the hard set. We can see that the model could utilize the structural information provided by the room layout to synthesize target-view panoramas with more visual-plausible layout structures.

\subsection{Panoramic View Synthesis on Real Datasets}

To verify the generalization ability of our method, we also conduct panoramic view synthesis on real datasets. We train the model on our dataset, then directly test it on 2D-3D-S dataset~\cite{ArmeniSZS2017} and PanoContext dataset~\cite{ZhangSTX14}. For each dataset, we set the camera translation distance as \SI{0.5}{\meter}, \SI{1.0}{\meter} and \SI{1.5}{\meter}, along the $x$-axis or $y$-axis randomly. \figref{fig:inference} shows the qualitative results. The results show that our method generalizes well to the real scenes and has a great potentiality for real-world application. To be noted, the vertical FoV of the panorama in 2D-3D-S dataset does not cover \SI{180}{\degree}. Thus, there are wavy black regions on the top and the bottom, making the synthesis more challenging and unrealistic to be inpainted completely.

\section{Conclusion}

In this paper, we explore synthesizing \SI{360}{\degree} novel views from a single indoor panorama and consider large camera translations. We propose a novel layout-guided method that exploits the room layout as a prior and geometric constraint. We also build a large-scale dataset for this novel task. The experiments show that our method achieves state-of-the-art performance and generalizes well on real datasets. In the future, we plan to exploit more general structures (\eg, planes or wireframes) and extend this idea to outdoor scenes.

\PAR{Acknowledgements.} This work was supported by the National Key R\&D Program of China (2018AAA0100704), the National Natural Science Foundation of China (61932020), Science and Technology Commission of Shanghai Municipality (20ZR1436000), and ``Shuguang Program'' by Shanghai Education Development Foundation and Shanghai Municipal Education Commission.

{
\small
\bibliographystyle{ieee_fullname}
\bibliography{references}
}

\end{document}